 \documentclass{ecai} 

\usepackage{latexsym}
\usepackage{amssymb}
\usepackage{amsmath}
\usepackage{amsthm}
\usepackage{booktabs}
\usepackage{enumitem}
\usepackage{graphicx}
\usepackage{color}
\usepackage{float}

\newcommand{\BibTeX}{B\kern-.05em{\sc i\kern-.025em b}\kern-.08em\TeX}

\begin{document}


\begin{frontmatter}


\paperid{123} 


\title{Improving LLMs for Machine Translation Using Synthetic Preference Data}


\author{\fnms{Dario}~\snm{Vajda}\thanks{Corresponding Author. Email: vajdadario@gmail.com}}
\author{\fnms{Domen}~\snm{Vreš}}
\author{\fnms{Marko}~\snm{Robnik-Šikonja}}
\address{University of Ljubljana, Faculty of Computer and Information Science}




\begin{abstract}
Large language models have emerged as effective machine translation systems. In this paper, we explore how a general instruction-tuned large language model can be improved for machine translation using relatively few easily produced data resources. Using Slovene as a use case, we improve the GaMS-9B-Instruct model using Direct Preference Optimization (DPO) training on a programmatically curated and enhanced subset of a public dataset. As DPO requires pairs of quality-ranked instances, we generated its training dataset by translating English Wikipedia articles using two LLMs, GaMS-9B-Instruct and EuroLLM-9B-Instruct. We ranked the resulting translations based on heuristics coupled with automatic evaluation metrics such as COMET. The evaluation shows that our fine-tuned model outperforms both models involved in the dataset generation. In comparison to the baseline models, the fine-tuned model achieved a COMET score gain of around 0.04 and 0.02, respectively, on translating Wikipedia articles. It also more consistently avoids language and formatting errors.
\end{abstract}

\end{frontmatter}


\section{Introduction}

Decoder-only large language models (LLMs) serve as versatile tools for a variety of natural language processing tasks, such as question answering, summarization, and translation. Typically, LLMs undergo three phases of training: pretraining, supervised fine-tuning (SFT), and preference alignment. The quality of a translation depends on many fine details (e.g., style, semantics, figurative language, etc.), which might not be sufficiently well learned during SFT. Our hypothesis is that a model can improve on subtle differences between a reasonable and good translation through preference alignment. In this work, we present the training of an LLM with Direct Preference Optimization (DPO) \citep{rafailov2024directpreferenceoptimizationlanguage} for optimizing its translation abilities. Using Slovene as a use case, our primary goal is to develop a reliable open-source English to Slovene translator that can be used for translating large English corpora to Slovene.

The main contributions of our research are:
\begin{itemize}
  \item A language agnostic method for improving translation models. The approach is based on synthetic data generation and is suitable for less-resourced languages such as Slovene.
  \item An open-source English to Slovene translation model capable of reliably generating high-quality translations.\footnote{\url{https://huggingface.co/GaMS-Beta/GaMS-DPO-Translator}}
  \item Source code for our data generation and fine-tuning pipeline for easier reproduction.\footnote{\url{https://github.com/DarioVajda/povejmo_dpo}}
\end{itemize}

Less-resourced languages, such as Slovene, lack sufficient high-quality data. One way to obtain more data is by translating high-quality English corpora. However, for Slovene, current open-source translators, such as RSDO~\cite{rsdo}, are not reliable enough for such a task, and the most successful commercial models, such as DeepL\footnote{\url{https://www.deepl.com}} are too expensive for translating large corpora. Hence, there is a need for a reliable open-source English to Slovene machine translation model.

Currently, there are no existing preference-annotated datasets for English to Slovene translation. Obtaining such a dataset using human translators and annotators would be slow and prohibitively expensive. Hence, we automatically create such a dataset. Our core insight is that even without access to a human-curated preference corpus for English to Slovene translation, we can bootstrap a reliable preference dataset by exploiting the behavior of two independent LLMs and a suite of automated filters. By prompting both GaMS-9B-Instruct \citep{gams9binstruct} and EuroLLM-9B-Instruct \citep{martins2025eurollm9btechnicalreport} to translate English Wikipedia articles~\cite{wikipedia} and a collection of English news articles from Common Crawl (CC-News dataset), we generate dual translations for each article, resulting in a dataset with around 67,000 entries from Wikipedia and 30,000 from CC-News. Whenever one model produces a clean Slovene output and the other makes an obvious error - whether by continuing the conversation in the wrong language, by truncating the translation, or by adding unwanted prefixes - we can confidently mark the former as \textit{chosen} and the latter as \textit{rejected}. To capture finer quality differences, we score all translations without any obvious mistakes with COMET \citep{rei2020cometneuralframeworkmt} and select pairs whose COMET scores differ by some minimum threshold. The result is a diverse, synthetic preference dataset that reflects both unacceptable errors (wrong language, incomplete output, etc.) and subtler language fluency distinctions captured by the COMET scores. After curating the translation pairs, our final preference dataset consists of around 25,000 entries for Wikipedia and 10,000 from CC-News.

Using generated synthetic preference data, we apply Direct Preference Optimization (DPO) to the GaMS-9B-Instruct model. We chose DPO because it directly optimizes the likelihood ratio between chosen and rejected outputs, sidestepping the instability and reward-modeling overhead of standard Reinforcement Learning from Human Feedback (RLHF) pipelines \citep{christiano2023deepreinforcementlearninghuman}. We train the model on the generated preference pairs over three epochs on 16 A100 GPUs. We employ a linear learning-rate warm-up followed by cosine decay to stabilize the early DPO updates. This fine-tuning framework takes advantage of both our filtering heuristics and DPO's principled ranking objective, driving the 9B-parameter model toward more fluent and complete Slovene translations.

We evaluate the effectiveness of our approach on the public SloBench leaderboard\footnote{\url{https://slobench.cjvt.si/}} and on unseen Wikipedia and CC-News articles. We show that our DPO-trained GaMS-9B-Instruct model outperforms the original model variant on SloBench and almost matches the performance of considerably larger GaMS-27B-Instruct \citep{gams27binstruct}. Additionally, on our Wikipedia benchmark, testing for language and formatting consistency, our model achieves an error rate of $ 0.8~\% $, which is a substantial improvement over the original model's $ 12~\% $.

The rest of the paper is organized into five sections. In Section \ref{sec:relatedwork}, we present related work on the development of Slovene language models and other tools we rely on. In Section \ref{sec:data}, we explain how the training data was generated through our heuristic approach and automatic metrics. The training pipeline and hyperparameter search are presented in Section \ref{sec:model}. In Section \ref{sec:evaluation}, we evaluate our model and compare it to other state-of-the-art open-source models. We provide conclusions, limitations, and directions for further work in Section \ref{sec:conclusion}. 

\section{Background and Related work}
\label{sec:relatedwork}
Our work is related to other research on LLMs for less-resourced languages, like Slovene, the ongoing research in preference alignment of LLMs, machine translation, and evaluation methods for machine translation. These topics are outlined below.

\subsection{Slovene Large Language Models}

Most current open-source LLMs are trained on predominantly English data. However, there are some multilingual models that support Slovene. Examples of such models are EuroLLM \citep{martins2025eurollm9btechnicalreport}, Gemma 2 \citep{Riviere2024Gemma2I}, and Gemma 3 \citep{gemmateam2025gemma3technicalreport}. These models were trained on multilingual datasets, with small amounts of Slovene texts. The portion of Slovene data in these datasets was relatively small (e.g., the EuroLLM  dataset consists of only around $1~\%$ Slovene texts), meaning there is room for improvement.

Efforts to develop strong LLMs for Slovene have primarily focused on adapting existing English-centric models due to the high cost of training an LLM from scratch. A notable early initiative in this area was the development of SlovenianGPT \citep{SlovenianGPT} and GaMS-1B \citep{GaMS-1B}. Those were followed by the development of 2B, 9B, and 27B parameter versions of the GaMS model. These developments demonstrated a key methodology for less-resourced languages: continuing the pre-training of a powerful base model on a relatively large corpus of Slovene text. The process also involved creating a new subword tokenizer adapted to the specifics of the Slovene language and employing embedding initialization techniques to transfer knowledge from the original English model.


\subsection{Machine Translators}
Inspired by Vaswani et al.~\citep{vaswani2023attentionneed}, machine translators have typically been built on encoder–decoder Transformer architectures. Later, OpenAI proved that decoder-only language models have the potential to learn many language-related tasks \citep{radford2019language}, including machine translation. They showed it by training their model for translation between English and French. 

The performance of machine translators on less-resourced languages such as Slovene often does not match the performance of models for high-resource languages. For example, a Slovene open-source RSDO Neural Machine Translator encoder-decoder model trained specifically for English-to-Slovene translation performs poorly on unseen domains compared to recent state-of-the-art models. Currently, the best-performing machine translators for Slovene, based on a public SloBench leaderboard, are DeepL, Claude, GPT, and Gemini. However, these models are commercial and translating larger corpora with them is costly. The best open-source models on this benchmark are EuroLLM and GaMS. However, as our preliminary evaluation shows, these models are unreliable for translating larger corpora, making trivial mistakes on some occasions. In our work, we focus on fixing such mistakes.

Multilingual open-source translators, such as NVIDIA Riva \citep{nvidia_riva_docs} and Meta’s No Language Left Behind (NLLB) \citep{NLLBTeam2024} perform worse than GaMS and EuroLLM on English-to-Slovene translation. NVIDIA Riva is a GPU-accelerated SDK (Software Development Kit) for building Speech AI applications, focusing on neural machine translation (NMT) while NLLB is a single, massively multilingual model that leverages transfer learning across languages to improve the translation quality of low-resourced languages.

\subsection{Preference-Based Model Alignment}
Preference alignment methods are used to improve the quality of large language models' outputs, and a few approaches have proven to be very effective at accomplishing that goal. The most classic example is Reinforcement Learning from Human Feedback (RLHF), that is a relatively complex process requiring training of a dedicated reward model. On the other hand, Direct Preference Optimization has recently gained significant traction and was shown to give competitive results without relying on an external reward model. With DPO the reward model is "embedded" inside the LLM and allows for a simpler and more efficient training pipeline with the goal of maximising the log-probability of chosen responses, $\log \pi_{\theta}(y_w|x)$, while minimising that of rejected responses, $\log \pi_{\theta}(y_l|x)$.


\subsection{Automatic MT Quality Metrics}
One of the standard automatic MT metrics is BLEU (Bilingual Evaluation Understudy) \citep{papineni-etal-2002-bleu}, which relies on n-gram overlap with reference translations and is incorporated into the SloBench evaluation. Recently, a shift towards learned metrics was driven by the need to capture semantic meaning, not just word overlap. The COMET framework uses cross-lingual embeddings to achieve a much higher correlation with human quality assessments. Therefore, to evaluate the translation pairs in our dataset, we employ the state-of-the-art reference-less Direct Assessment (DA) model from CometKiwi \citep{cometkiwi} known for its high correlation with human judgment.

\section{Synthetic Preference Data Generation}
\label{sec:data}

Obtaining a high-quality preference annotated translation dataset is challenging. We take English Wikipedia and news articles from Common Crawl as a starting point, as those cover a wide variety of topics. Our data generation pipeline consists of multiple stages. We start by generating translations, described in Section \ref{sec:translations}. This is followed by identifying trivial errors, described in Section \ref{sec:failure_modes}, and scoring remaining instances, described in Section \ref{sec:scoring}. Our final data construction is described in Section \ref{sec:final_data}.

\subsection{Generating Pairs of Translations}
\label{sec:translations}

The main challenge of generating a synthetic preference dataset is generating distinct translation candidates. Notably, the generated errors shall not propagate or accumulate through the process. Our approach is to generate candidate translations from a corpus of articles which have been selected that cover a broad range of topics with two distinct models and rank the responses using automatic quality metrics.

The first model we use is GaMS-9B-Instruct~\cite{gams9binstruct}, which is based on the Gemma 2 architecture and adapted for Slovene. As this is the model we also aim to fine-tune as the final machine translator, this allows us to construct preference pairs that target the model's natural output distribution. The second model we utilize for generating the translations is EuroLLM-9B-Instruct~\cite{martins2025eurollm9btechnicalreport}. We chose this model because it is an open-source model that demonstrated strong performance and reliability for English-to-Slovene translation in our preliminary experiments.

We use these two models to translate over 67,000 Wikipedia articles, consisting of approximately 26 million words. We use the following prompt instructions:
\begin{itemize}
    \item GaMS: \textit{"Prevedi naslednje angleško besedilo v slovenščino."} (en. \textit{"Translate the following English text to Slovenian."})
    \item Euro-LLM: \textit{"Translate the following English text to Slovenian."}
\end{itemize}
We filter the initial pool of translations during the subsequent data curation steps.

\subsection{Identifying Failure Modes}
\label{sec:failure_modes}

Upon inspecting the translations, we identified several error types that are critical to the model's reliability, yet simple to represent as preference pairs with unambiguous chosen and rejected examples.

The most significant failure mode observed was generating outputs in the wrong language. To programmatically verify the language of each translation, we utilize the pre-trained language identification model from the FastText library \citep{joulin2016fasttext, joulin2016bag}. We use a lightweight and efficient classifier capable of accurately identifying 176 different languages from raw text, making it highly suitable for large-scale filtering tasks. This process forms high-confidence preference pairs by identifying instances where one generated translation is in Slovene and the other is in a different language. The correct, Slovene translation is labeled as \textit{chosen}, while the incorrect one is labeled as \textit{rejected}.

Another identified failure mode is translation truncation, where the model only translates a portion of the source text. We hypothesize that this behavior with GaMS-9B-Instruct is a result of its SFT dataset containing only sentence-level translation tasks. Therefore, the model learned to respond to translation tasks with short answers. This type of structural error is particularly well-suited for correction with DPO. To address this, we create preference pairs from instances where both translations are in Slovene, but one is complete while the other is clearly truncated. The complete translation is labeled as \textit{chosen}, and the truncated version as \textit{rejected}. A translation was considered truncated with high confidence if it was less than 50\% of the length of the original text, measured by character count.

A more subtle issue we identified was the presence of stylistic formatting artifacts. Sometimes the model starts a response with \textit{"Slovenski prevod:"} (en. Slovene translation), \textit{"Slovene translation:"}, etc. Since the goal is to produce only the translated text, this behavior is addressed by creating a specific type of preference pairs for our training dataset. Translation pairs were constructed in the following manner: the \textit{chosen} response is a clean, complete translation, while the corresponding \textit{rejected} response is created by prepending the \textit{chosen} text with one of the undesirable prefixes. This method provides a clear and direct preference signal to the model during DPO.

\subsection{Scoring and Filtering the Translations}
\label{sec:scoring}

While the initial heuristic filtering addresses clear structural errors, discerning finer differences in quality requires a quantitative metric. For this purpose, we employ the COMET score, specifically the Unbabel/wmt22-cometkiwi-da model \cite{cometkiwi}. We select this model as it is a state-of-the-art, reference-less Direct Assessment (DA) model that excels at predicting translation quality with a high degree of correlation to human judgment.

All translation pairs that pass the initial heuristic checks are then scored using this COMET model. The scores serves as a proxy for human preference. Since many translation pairs exhibit only minor quality differences, we introduce a minimal score difference threshold to take a given translation pair into consideration. This step is crucial to prevent metric noise from being misinterpreted as a meaningful preference signal. Consequently, a preference pair is only created if the absolute score difference between the two candidates is greater than 0.05. The translation with the higher score is labeled as \textit{chosen} and the other as \textit{rejected}.

\subsection{Constructing the preference dataset}
\label{sec:final_data}

The preference pairs generated from the preceding heuristic and metric-based methods are merged to form the final training dataset. This combined dataset is designed to capture both critical failure modes of the base model as well as more subtle preference signals based on clarity, grammar, and style. The above process reduces the number of translation pairs from a total of 107,000 to approximately 35,000. The number of translation pairs for our dataset was decreased because not all of them carried useful information. Additionally, the synthetically generated formatting pair count was chosen to make up around 20\% of the final dataset. Those were added to the other systematically curated pairs from the original translation pair dataset. The distribution of training examples is as follows:
\begin{itemize}
  \item Pairs targeting incorrect language - $22~\%$
  \item Pairs targeting response truncation - $3~\%$
  \item Synthetically generated formatting pairs - $20~\%$
  \item Pairs derived from COMET score differences - $55~\%$
\end{itemize}

\section{Translator training}
\label{sec:model}

To produce an improved LLM for MT, we take GaMS-9B-Instruct model as our starting point and optimize it for translation on the preference dataset from Section~\ref{sec:data} using the DPO method. We provide a brief description of the method in Section \ref{sec:dpo}. We describe our training implementation in Section~\ref{sec:implementation}. In Section~\ref{sec:hyperparameters}, we describe the hyperparameter search performed.

\subsection{Using DPO for Machine Translation}
\label{sec:dpo}

Traditional approaches for preference alignment, such as RLHF, rely on training a dedicated reward model. Since we are using a synthetic preference dataset, using another synthetic data-based model in our training pipeline would introduce additional noise and risk instability in the fine-tuning process. 

Therefore, we chose Direct Preference Optimization (DPO) as our fine-tuning method due to its stability and efficiency compared to traditional reinforcement learning-based approaches like Proximal Policy Optimization (PPO). DPO provides a cleaner and more straightforward training pipeline with less room for error accumulation. As the DPO loss function is mathematically equivalent to the objective in traditional RLHF, it offers the same optimization guarantees within a more direct and stable framework. Given the data distribution $ \mathcal{D} = \{(x, y_w, y_l)\} $, where $x$ is the model's input, $ y_w $ denotes the chosen (preferred) response and $ y_l $ denotes the rejected response, the DPO loss function aims to minimise:
\begin{equation}
\begin{split}
\mathcal{L}_{\text{DPO}}(\pi_{\theta}; \pi_{\text{ref}}) = & -\mathbb{E}_{(x, y_w, y_l) \sim \mathcal{D}} \Biggl[ \log \sigma \Biggl( \beta \log \frac{\pi_{\theta}(y_w | x)}{\pi_{\text{ref}}(y_w | x)} \\
& \qquad - \beta \log \frac{\pi_{\theta}(y_l | x)}{\pi_{\text{ref}}(y_l | x)} \Biggr) \Biggr]
\end{split}
\label{eq:dpo_loss}
\end{equation}
In this formulation, $\pi_{\theta}$ represents the fine-tuned policy (model) and $\pi_{\text{ref}}$ denotes the reference policy (usually the starting model). The reference model, $\pi_{\text{ref}}$, is a crucial component that regularizes the training process. For our experiments, we use the initial GaMS-9B-Instruct model as the reference model. The temperature parameter, $\beta$, controls how strongly the policy model adheres to the preference data. A higher $\beta$ leads to a closer fit to the preference pairs, while a lower $\beta$ maintains closer proximity to the reference model's initial behavior. We determine the exact $\beta$ value through hyperparameter tuning.

\subsection{Implementation and Training Environment}
\label{sec:implementation}

We use the HuggingFace Transformers \citep{wolf-etal-2020-transformers} DPO implementation. Specifically, we use the \texttt{TRL} (Transformers Reinforcement Learning) \citep{vonwerra2022trl} library in combination with \texttt{Accelerate} \citep{accelerate} and \texttt{Deepspeed} libraries.

To make training of a 9B-parameter model computationally feasible, we employ a parameter-efficient approach. Specifically, we use Low-Rank Adaptation (LoRA) \citep{hu2021loralowrankadaptationlarge} from the \texttt{peft} \citep{peft} library. LoRA enables efficient adaptation of large pre-trained models by introducing and training only low‐rank update matrices, thereby reducing the number of trainable parameters by orders of magnitude, lowering both GPU memory and storage requirements. This parameter-efficient approach accelerates fine-tuning and simplifies model deployment, while achieving comparable task performance to full model fine-tuning.

We performed the training on the Slovene HPC Vega supercomputer. We utilized a configuration of 4 compute nodes, each equipped with 4 NVIDIA A100 40GB GPUs, for a total of 16 GPUs per training run. The GPUs on a single node are connected using NVLINK with total bandwidth of 600 GB/s. The nodes are connected through 2x200 Gb/s InfiniBand switches in Dragonfly+ topology. 

To manage the substantial memory requirements of fine-tuning the 9B-parameter model, even with LoRA, we employed the ZeRO (Zero Redundancy Optimizer) Stage 2 optimization strategy \citep{rajbhandari2020zeromemoryoptimizationstraining}, as implemented in the DeepSpeed library. This technique mitigates memory redundancy across data-parallel workers by partitioning not only the optimizer states (as in Stage 1) but also the gradients. While each GPU maintains a complete copy of the model's parameters for the forward and backward passes, it is only responsible for storing and updating a distinct shard of the gradients and corresponding optimizer states. After the local optimizer step, an all-gather operation efficiently synchronizes the updated weights across all GPUs, ensuring model consistency for the next iteration. This approach dramatically reduces the per-GPU memory footprint compared to standard data parallelism, making it feasible to fine-tune the full model on our 16 A100 GPU setup without resorting to more complex model parallelism.

The final major optimization we use is Gradient Checkpointing \citep{chen2016trainingdeepnetssublinear}. It trades additional computation (around 30–40\% increase for LLMs) for reduced activation and gradient memory usage (by a factor of approximately $\sqrt{\text{num\_layers}}$) by selectively storing only a subset of intermediate activations during the forward pass and recomputing the omitted activations \textit{on-the-fly} in the backward pass. This enables training of much deeper or wider models under fixed memory budgets, making it particularly valuable for large-scale deep training or fine-tuning.

\subsection{Training and Hyperparameter Grid-Search}
\label{sec:hyperparameters}

We split the curated preference dataset only with translations of Wikipedia articles into the training and validation sets with 24,000 and 1,000 instances, respectively. The validation set was used to monitor performance during training and to select the optimal hyperparameters through a grid search.

\begin{figure}[htb]
  \centering
  \includegraphics[width=\columnwidth]{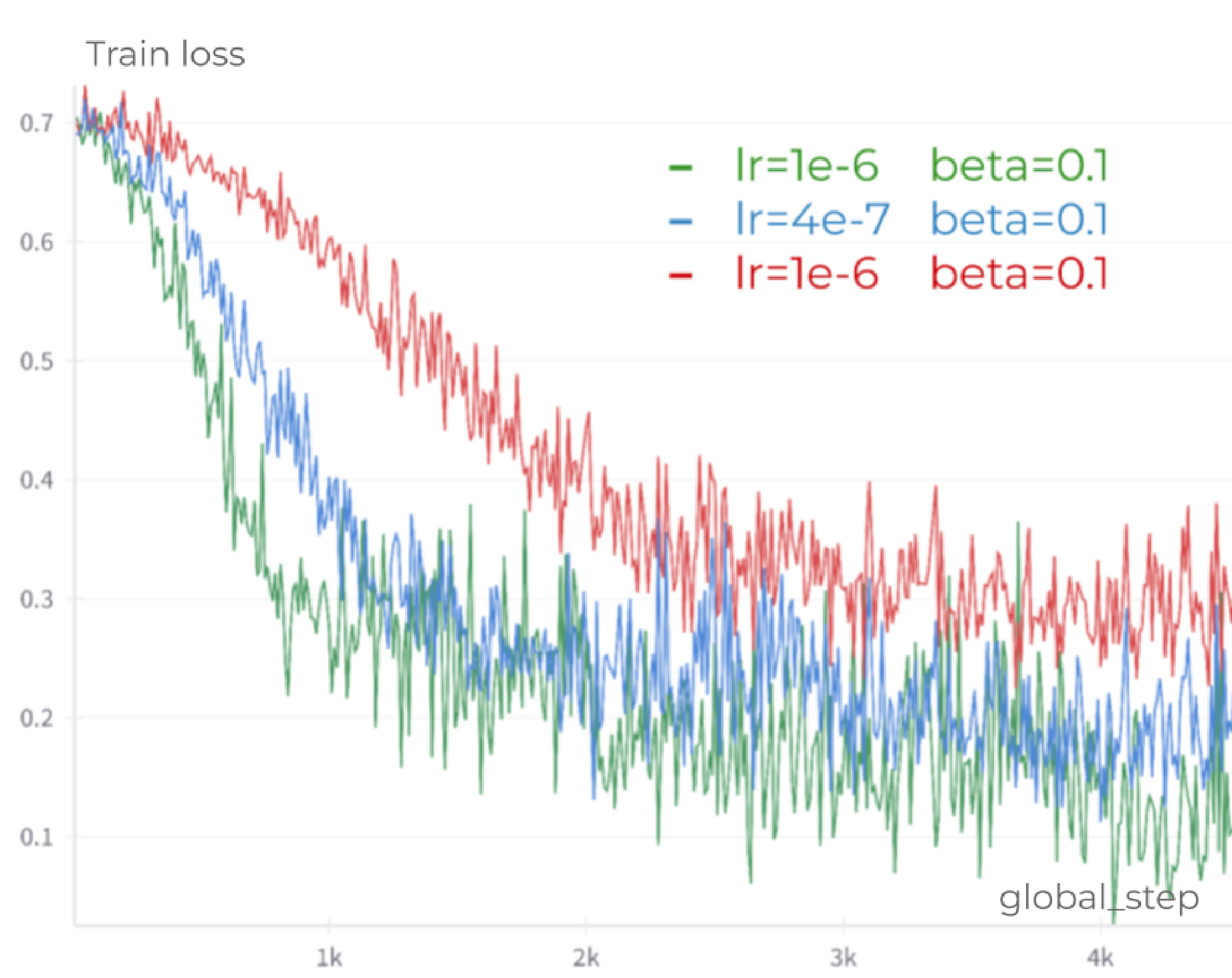}
  \includegraphics[width=\columnwidth]{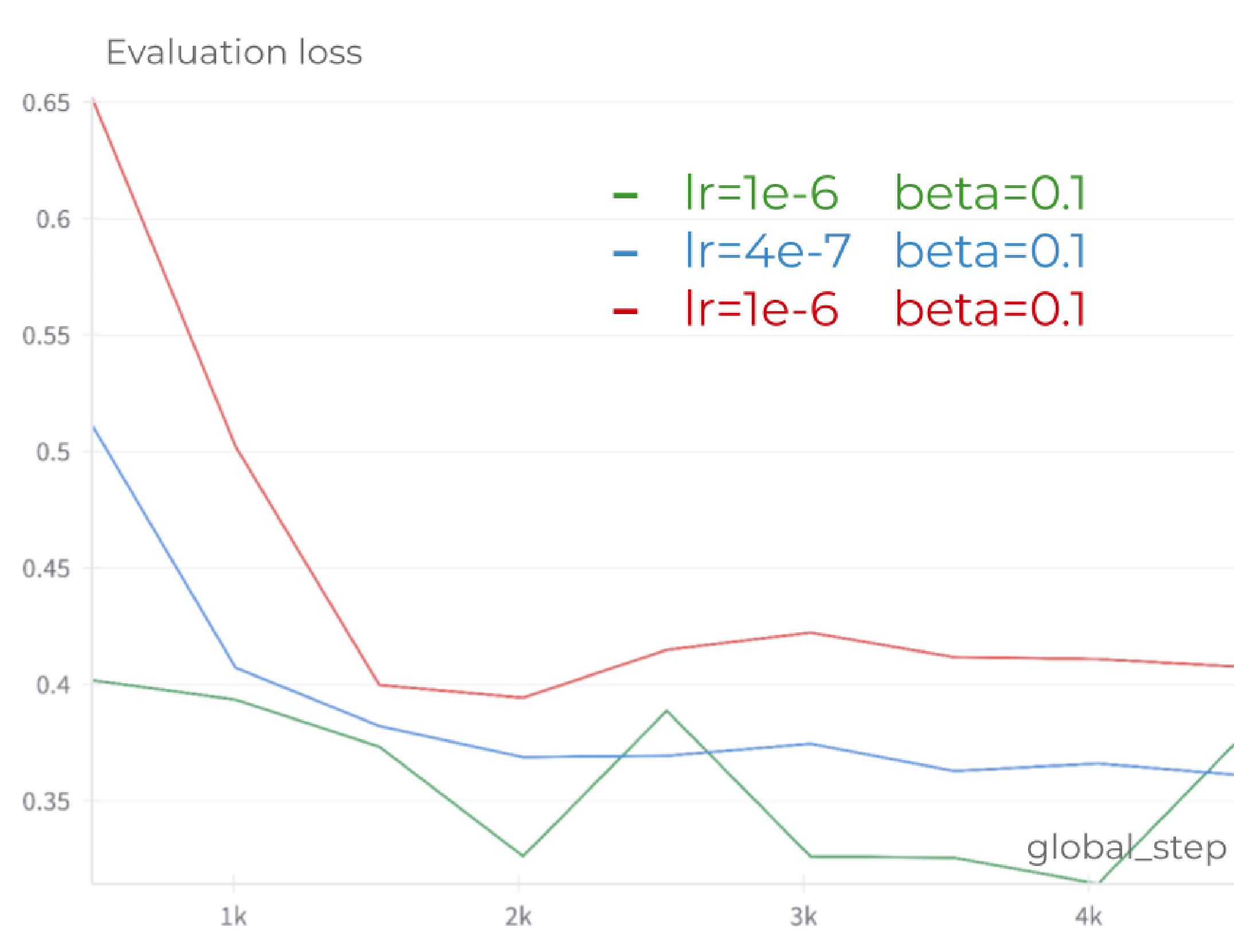}
  \caption{DPO training (top) and validation loss (bottom) curves for different learning rate and DPO $\beta$ hyperparameter values.}
  \label{fig:loss}
\end{figure}

The key hyperparameters for our DPO training runs are detailed in Table~\ref{tab:train-hparams}. We conducted a grid search over the DPO $\beta$ and learning rate. The final model was trained for 3 epochs using the optimal configuration discovered during this search. To prevent overfitting, we compare all checkpoints using validation loss. Observed training and validation losses are shown in Figure~\ref{fig:loss}. Since we achieved the lowest validation loss of $0.315$ for hyperparameter values $\beta=0.1$ and $lr=1\cdot10^{-6}$ at the second-to-last evaluation step, this is the final version of our model. Each one of the grid-search training runs lasted around 5 hours. 

\begin{table}[htb]
  \centering
  \caption{Training hyperparameter values. For DPO $\beta$ and learning rate, the search domains are provided. The bold values were selected as optimal.}
  \label{tab:train-hparams}
  \begin{tabular}{@{} ll @{}}
    \toprule
    Parameter        & Value \\ 
    \midrule
    Epochs              & 3 \\
    Micro batch size    & 1 \\
    Global batch size   & 16 \\
    \textbf{DPO} $\mathbf{\beta}$            & \{\textbf{0.1}, 0.2\} \\
    LoRA rank           & 64 \\
    \textbf{Learning rate}       & \{\textbf{1e-6}, 4e-7, 1e-7\} \\
    Warmup steps        & 1500 \\
    Learning rate scheduler & cosine\_with\_min\_lr \\
    \bottomrule
  \end{tabular}
\end{table}

Once the optimal hyperparameter configuration was found, the training dataset was expanded by adding the translation pairs from CC-News (approximately 10,000 new training examples). The model was trained on the larger dataset for three epochs. This final fine-tuning on the complete dataset with the given hyperparameters took approximately 7 hours. This training run resulted in a model checkpoint with a noticeably lower validation loss of $0.255$. This checkpoint is the model we will be evaluating further and comparing it to GaMS-9B-Instruct, EuroLLM-9B-Instruct and some other models.

\section{Evaluation and Results}
\label{sec:evaluation}

We evaluate our model on two benchmarks. The first is a public Slovene-to-English benchmark, that is part of the public benchmark suite SloBench. Since SloBench evaluates the models only on per-sentence translations, we also evaluate our model on a custom evaluation set relevant to our specific goal of translating longer English documents for training large language models for Slovene. We create such an evaluation set based on a set of unseen English Wikipedia and CC-News articles.
Throughout this section, we refer to our model as GaMS-9B-DPO-Translator.

\subsection{SloBench Evaluation}

SloBench is an evaluation platform for benchmarking Slovene large language models and their capabilities. Since the benchmarks' ground truths are not public, we believe that benchmark tuning, leading to misleading results, is not possible, making this benchmark an objective measure for many Slovene natural language processing tasks.
We test the model on the \texttt{Machine Translation (ENG -> SLO)}\footnote{\url{https://slobench.cjvt.si/leaderboard/view/8}} task. The task consists of five different domains: Scientific articles, Speech texts, Legal articles, News articles, and Technical texts.

The results are shown in Table~\ref{tab:slobench}. Even though our model was not specifically tuned for any of the benchmark's domains, our DPO training resulted in a noticeable improvement in comparison to the base model (GaMS-9B-Instruct). GaMS-9B-DPO-Translator achieved a similar score to GaMS-27B-Instruct, which is three times larger.

The improvement would likely be even higher if our preference dataset included a broader range of data, not just Wikipedia documents. Therefore, incorporating more conversational, legal, and news texts into our data generation pipeline and training the model with DPO on such a dataset would better capture all domains tested by SloBench and potentially increase its score. Writing styles in those domains often differ from the Wikipedia articles we used.

\subsection{Wikipedia and CC-News Evaluation}
A more suitable evaluation method for our goal of translating large amounts of longer documents is to compare our fine-tuned model to the base model when translating English Wikipedia and CC-News articles that were not seen during training or validation.

The creation of an evaluation dataset was very similar to generating the training dataset. We translated 500 randomly chosen articles from both sources with GaMS-9B-Instruct (base model), GaMS-9B-DPO-Translator (fine-tuned model), and EuroLLM-9B-Instruct (for reference). Those articles were chosen in a way that none of those have been seen during training or validation by our fine-tuned model.

The first step in analyzing the performance of our model  is to check for any of the trivial mistakes we already uncovered when preparing the training dataset. The comparison of models on such mistakes is shown in Table~\ref{tab:wikierrors}.

\begin{table}[H]
  \centering
  \caption{Error rates comparison on our custom Wikipedia Evaluation dataset. Each model name refers to its 9B parameter instruction-tuned variant. }
  \label{tab:wikierrors}
  \begin{tabular}{lccc}
    \toprule
    \textbf{Model} & \textbf{Language} & \textbf{Truncation} & \textbf{Combined} \\
    \textbf{} & \textbf{Error} & \textbf{Error} & \textbf{} \\
    \midrule
    EuroLLM-9B-Instruct & 1\%   & 0.4\% & 1.4\% \\
    GaMS-9B-Instruct    & 9.5\% & 3.5\% & 13\% \\
    \textbf{GaMS-9B-DPO-Translator}   & \textbf{0.6\%} & \textbf{0.2\%} & \textbf{0.8\%} \\
    \bottomrule
  \end{tabular}
\end{table}

For the articles without trivial errors, we calculated the COMET scores of each model's translation and compared them. To ensure fair comparison, only those articles were used where none of the models made any critical mistakes. Obtained COMET scores are shown in Table \ref{tab:wikicomet}.

\begin{table}[H]
  \centering
  \caption{Comparison of COMET scores on our custom Wikipedia Evaluation dataset. Higher score is better.  }
  \label{tab:wikicomet}
  \begin{tabular}{lccc}
    \toprule
    \textbf{Model} & \textbf{Wikipedia} & \textbf{CC-News} &\textbf{Average} \\
    \midrule
    EuroLLM-9B-Instruct              & 0.727 & 0.667 & 0.695 \\
    GaMS-9B-Instruct                 & 0.722 & 0.680 & 0.698 \\
    \textbf{GaMS-9B-DPO-Translator}  & \textbf{0.757} & \textbf{0.715} & \textbf{0.735} \\
    \bottomrule
  \end{tabular}
\end{table}

\begin{table*}[!htb]
  \centering
  \caption{Comparison of different Machine translation models on the public SloBench English-to-Slovene translation leaderboard. The results of OpenAI GPT 4o-mini, GaMS-9B-Instruct, GaMS-27B-Instruct, and EuroLLM-9B-Instruct are taken directly from the leaderboard. The results of our model are in bold. }
  \label{tab:slobench}
  \resizebox{\textwidth}{!}{%
    \begin{tabular}{l|cccccc}
      \toprule
      \textbf{Model}
        & \textbf{BERT score}
        & \textbf{BLEU (avg)}
        & \textbf{METEOR (avg)}
        & \textbf{CHRF (avg)}
        & \textbf{BLEU (corpus)}
        & \textbf{CHRF (corpus)} \\
      \midrule
      EuroLLM-9B-Instruct        & 0.8741 & 0.2927 & 0.5792 & 0.6055 & 0.3273 & 0.6055 \\
      GaMS-27B-Instruct          & 0.8734 & 0.2866 & 0.5688 & 0.5986 & 0.3246 & 0.5986 \\
      \textbf{GaMS-9B-DPO-Translator}
                                 & \textbf{0.8726}
                                 & \textbf{0.2810}
                                 & \textbf{0.5663}
                                 & \textbf{0.5967}
                                 & \textbf{0.3252}
                                 & \textbf{0.5967} \\
      GaMS-9B-Instruct           & 0.8713 & 0.2773 & 0.5616 & 0.5928 & 0.3209 & 0.5928 \\
      GPT 4o-mini                & 0.8690 & 0.2619 & 0.5456 & 0.5839 & 0.3021 & 0.5839 \\
      \bottomrule
    \end{tabular}%
  }
\end{table*}

Our fine-tuned model outperformed both models that were used in the dataset construction, showing that those models do not necessarily represent an upper limit for fine-tuning performance. The reason why our fine-tuned model is able to outperform the construction models is that DPO does not directly train the model to replicate all translations from the dataset. This allowed our model to learn good aspects of both EuroLLM-9B-Instruct and GaMS-9B-Instruct, but also helped it to avoid their mistakes.

There is a discrepancy between the results on SloBench and the results on our custom Wikipedia article translation test when we compare EuroLLM-9B-Instruct and our fine-tuned GaMS-9B-DPO-Translator. We hypothesize that this is due to the difference in example lengths between benchmarks. The model should benefit from our training, especially on longer texts, as the errors (wrong language, truncation) of the base model were rarer on shorter texts. 

A limitation of our evaluation on Wikipedia articles is that it exhibits a similar distribution to our training data, making our model more likely to perform well, and those circumstances might have benefited our model in comparison to others. However,  this limitation does not dispute the fact that we successfully improved the model and achieved our initial goal of reliably and accurately translating longer documents. Since Wikipedia captures a variety of different fields and topics, this learned knowledge should carry over to other types of documents, which will be useful for generating new training data for Slovene LLMs.

\subsection{Efficiency of our solution}
Our solution is very efficient from the training data acquisition standpoint and eliminates the need for manual labeling, since it doesn't require any human annotators. 

\paragraph{Data acquisition.}
The dataset creation pipeline is computationally efficient since it involves batched inference of LLMs and other lightweight models such as the COMET model for translation scoring and the language identification model from the FastText library. This process overcomes the challenge of obtaining high quality training data for low-resource languages, required by SFT. In our case, to acquire the dataset we used approximately 3 hours on one node with 4 A100 40GB GPUs (\textbf{12 GPU hours}).

\paragraph{Fine-tuning.}
On the other hand, fine-tuning the model is comparable to SFT, with a notable difference being that DPO requires two forward passes per example (for \textit{chosen} and \textit{rejected} responses). The fine-tuning was run on 4 nodes with 4 A100 40GB GPUs each and ran for around 7 hours (\textbf{112 GPU hours}). We used 4 nodes to speed up the process, but the minimum hardware requirement with a 9B parameter model for this step is only one such node and it should run for less than 28 hours (since computation time decreases at a close to linear rate with respect to the number of GPUs when using ZeRO stage 2), which is quite reasonable for a model like the one we used.

\section{Conclusion}
\label{sec:conclusion}

We proposed a pipeline for improving machine translation based on data generation and DPO preference alignment method. We showed that our approach increases the quality of the trained model's translations. We showed a small performance improvement on the SloBench evaluation, and a substantial improvement in translating longer documents, such as Wikipedia articles.

The main goal of our research was to make an open-source translator for a less-resourced language (Slovene) more reliable. Since our approach is language agnostic (given the precondition of the existence of at least two machine translation options for this language), it can be applied to many less-resourced languages or specific domains. We believe that our approach will help translate high-quality English corpora to less-resourced languages which is necessary to build LLMs in such languages and important for the sovereignty of such languages in the LLM era.

We plan to use the insight gained during this project to fine-tune the 27B parameter model with the same training pipeline. Since the systems are already in place, the remaining challenges are to scale the training process and to obtain the required computational resources. Scaling for a larger model would involve generating more training data and using more advanced distributed training optimizations such as ZeRO Stage 3, ZeRO++~\cite{wang2023zeroextremelyefficientcollective}. Additionally, the recently released NVIDIA NeMo-RL framework\footnote{\url{https://github.com/NVIDIA-NeMo/RL}} shall be tested.

A potential improvement to consider in the future is Curriculum DPO \citep{croitoru2025curriculumdirectpreferenceoptimization, pattnaik2024currydpoenhancingalignmentusing} instead of vanilla DPO. Curriculum learning would allow the model to learn on different datasets, step-by-step, increasing in difficulty. The datasets could be divided into two major groups. The first group would contain the training examples from our heuristic-based analysis (language and truncation errors), and the second group would have the training examples ranked by COMET score. The latter could be further subdivided into multiple subsets based on the COMET score difference between \textit{chosen} and \textit{rejected} responses. A lower score delta indicates a more subtle difference in quality, and the model would be trained on those after it had been trained on the pairs with a more obvious quality difference.

Finally, other preference alignment methods, such as GRPO, could be tested. Since we have already automated response rankings, we would have to turn those rankings into a reward function for GRPO. Another possibility would be combining both methods by first focusing on language and truncation errors using DPO and then performing GRPO based on COMET scores.









 \begin{ack}
The work was primarily supported by the EU cofinancing for research innovation projects in support of green transition and digitalisation (project PoVeJMo, no. C3.K8.IB). Further support was provided by the Slovene Research and Innovation Agency (ARIS) project GC-0002 and the core research programme P6-0411.  The work was also supported by EU through ERA Chair grant no. 101186647 (AI4DH). The computational resources were provided by SLING through project S24O01-42.
\end{ack}



\bibliography{mybibfile}

\end{document}